# Coronavirus disease situation analysis and prediction using machine learning: a study on Bangladeshi population

**Al-Akhir Nayan[1], Boonserm Kijsirikul[1], Yuji Iwahori[2]**
[1]Department of Computer Engineering, Faculty of Engineering, Chulalongkorn University, Bangkok, Thailand
[2]Department of Computer Science, College of Engineering, Chubu University, Aichi, Japan

**ABSTRACT**

During a pandemic, early prognostication of patient infected rates can reduce the death by ensuring treatment facility and proper resource allocation. In recent months, the number of death and infected rates has increased more distinguished than before in Bangladesh. The country is struggling to provide moderate medical treatment to many patients. This study distinguishes machine learning models and creates a prediction system to anticipate the infected and death rate for the coming days. Equipping a dataset with data from March 1, 2020, to August 10, 2021, a multi-layer perceptron (MLP) model was trained. The data was managed from a trusted government website and concocted manually for training purposes. Several test cases determine the model's accuracy and prediction capability. The comparison between specific models assumes that the MLP model has more reliable prediction capability than the support vector regression (SVR) and linear regression model. The model presents a report about the risky situation and impending coronavirus disease (COVID-19) attack. According to the prediction produced by the model, Bangladesh may suffer another COVID-19 attack, where the number of infected cases can be between 929 to 2443 and death cases between 19 to 57.

*Corresponding Author:*

Al-Akhir Nayan
Department of Computer Engineering, Faculty of Engineering, Chulalongkorn University
254 Phaya Thai Rd, Wang Mai, Pathum Wan District, Bangkok 10330, Thailand
Email: asquiren@gmail.com

## 1. INTRODUCTION

In December 2019, the coronavirus disease (COVID-19) first appeared in China [1]. Over 225.62 million cases were documented worldwide until September 2021, when the mortality rate was 2% [2]. From March 2020 to August 2021, more than 1.53 million people were infected with COVID-19, and 26,972 deaths were filed in Bangladesh, where the death rate was 1.79% [3], [4]. From June 2021 to August 2021, the infected and death rate crossed all previous records where the infected rate of several districts of Bangladesh was recorded to 60% [5], [6]. This accelerated pandemic spread in Bangladesh became a critical issue and a severe threat to public health and the economy. Most countries restricted social interaction by imposing isolation and quarantine to prevent the spreading virus. However, due to late identification of the unusual and unknown nature of the delta variant, many infected people did not get proper treatment.

Many researchers established novel approaches for screening infected individuals at various stages to uncover noticeable connections between clinical variables and the likelihood of succumbing to the disease. In recent research investigations, artificial intelligence (AI) [7]–[9] and machine learning (ML) [10]–[12] approaches have been found to have a vital role in minimizing the impact of viral dissemination. The use of ML on patient data is being studied in various ways. The essential research fields are classifying patients based on their clinical results and predicting infection and mortality rates [13], [14]. These research studies are critical, but the studies tremendously aid persons working in the health sector to be prepared adequately and take all essential safeguards to prevent a pandemic from spreading [15].

Previous studies on COVID-19 show unsatisfactory performance in analyzing and predicting Bangladesh's current situation. The insufficient data of the current situation, rapid spreading of COVID-19 among people, and the drawbacks of the used techniques are the major issues behind the lower accuracy of previous works. This research intends to improve a machine learning model to interpret and forecast Bangladesh's COVID-19 death and infected rate solving issues of previous works. The study distinguishes the prediction efficiency of well-established machine learning techniques. A dataset has been developed collecting patient data from the Bangladeshi government website [16] between March 1, 2020, and August 10, 2021. The results of the research findings have been explained elaborately using essential diagrams. The prophecy by the multi-layer perceptron (MLP) model has an excellent match with the actual death and infected rate of Bangladesh. Such a type of analysis and forecast model can be an efficient solution for Bangladesh to struggle against COVID-19. The code and dataset are uploaded to GitHub's private directory. After publication, the directory will be made public.

In the article, related works have been mentioned in section 2. The MLP model's architecture, working procedure of support vector regression (SVR) and linear regression model, data processing techniques, dataset configuration, training, and testing arguments have been described elaborately in section 3. Section 4 explains the infected and death case prediction, critical situation analysis, and comparison among the prediction accuracy by several related models. Lastly, a conclusion part has been added in section 5.

## 2. RELATED WORKS

Researchers in the computer science field are striving to forecast possible scenarios of the following days to assist the government in planning for the most egregiously wrong scenarios and acting as needed by allocating the right amount of assets. Many researchers have worked on COVID-19 using machine learning techniques, prioritizing the infected and death rate. The significance of the forecasting model in predicting the pandemic situation was described by Shinde *et al* [17]. They highlighted various criteria that play an essential role in pandemic conditions. They made some forecasting recommendations as part of their research. They proposed the most crucial variables for anticipating pandemic conditions. Their work recommended using machine learning models such as the Hill equation, the Weibull equation, and logistic regression to determine infection rates. They demonstrated that monitoring people is challenging, and forecasting models face challenges due to the long incubation period. The findings revealed that a lack of adequate data harms predictions [17].

Gupta *et al.* [20] investigated the infected case growth rate and discovered that lockdown events decrease the occurrence of infected cases. They employed polynomial regression and exponential regression models [18]. The least absolute shrinkage and selection operator, support vector machine (SVM), exponential smoothing, and linear regression were utilized by Rustam *et al.* to anticipate COVID-19 death cases recoveries and confirmed cases over the next ten days [19]. Abdullha and Abujar tried to analyze day-by-day data to know the COVID-19 situation. They utilized machine learning supervised models such as k-nearest neighbors (KNN) linear regression (LR) for analysis and prediction. They predicted death, infected, and recovery cases for 90 days and concluded that the KNN and LR algorithms maintain approximately similar prediction accuracy [20].

Satu *et al.* [21] argued that the COVID-19 situation has become more critical in Bangladesh due to a lack of effective treatment and necessary tools for analyzing and predicting the pandemic. They proposed a short-term prediction generating machine learning model based on the cloud. They trained a linear regression model with 25 days of data, generated seven days prediction, and identified factors responsible for COVID-19 [21].

Aljameel *et al.* [22] proposed a prediction method to identify COVID-19 earlier by monitoring the characteristics of patients at home. They worked on 287 samples and analyzed data using random forest (RF), logistic regression (LR), and extreme gradient boosting (XGB). Before analyzing, they partitioned data using 10 k-cross validation and fixed data imbalance using synthetic minority over-sampling technique (SMOTE). They concluded that RF performed better than the other two algorithms [22].

Hossain *et al.* [23] worked on research to identify factors responsible for COVID-19. They analyzed the impact of COVID-19 using the binary logistic regression model. Using the autoregressive integrated moving average (ARIMA) model, they predicted the number of confirmed and death cases for the next 40 days applying the autoregressive integrated moving average (ARIMA) model. Their prediction report found that confirmed cases decreased and death cases were constant. After analyzing different people, they concluded that education is essential to prevent COVID-19 [23].

Most of the studies used LR, logistic regression, extreme gradient boosting (XGB), RF, or KNN for analyzing and predicting the mortality and infected rate. But the research investigations revealed inadequate performance to examine and predict the present situation of Bangladesh. Some approaches are finishing with 30-40% prediction efficiency, which is not sufficient to forecast the contemporary situation of Bangladesh. The previous model's hindrances and inadequate performance paved the way for this research. In this work, we have employed the MLP model to investigate and forecast the situation on Bangladeshi perspective data between March 2020 and August 2021.

## 3. RESEARCH METHOD
### 3.1. The MLP model structure

MLP is the most prevalent neural network model in deep learning. Being more modest than other complex models, MLP paved the path for advanced neural networks [24]. The technique incorporates linked neurons that pass information back and forth, assigning a value to each neuron-like the human brain. The input layer acquires input, and the output layer performs prediction and categorization. The actual computational engine of the model is an arbitrary number of hidden layers embedded between the input and output layers. A feedforward neural network transmits data in the forward direction from the input layer to the output layer assigning weights to establish links between the layers. The effect of a connection is determined by its weight which is the main principle of an MLP's learning process [25], [26]. An MLP has been modeled utilizing a hundred hidden layers, each with 64 neurons, and each node associates to all upper and lower nodes. The model contains 102 layers, including the input and output layers. 'Adam' is used as the default optimizer by MLP but to handle the large dataset, 'lbfgs' optimizer has substituted the default one. Figure 1 shows the schematic representation of the modified MLP layers.

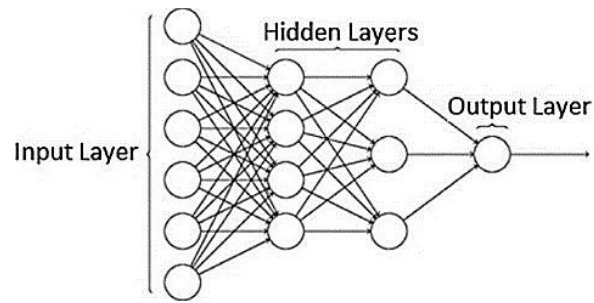

Figure 1. MLP layer's schematic representation

### 3.2. MLP weights optimization

A backpropagation technique is utilized to optimize the model's weights that receive the outputs as inputs. Random weights are assigned to all connections in a conventional MLP. These randomized weights transfer values throughout the network, resulting in the final output. This output would deviate from the predicted result, and the error is the difference between the two values.

According to the process, an error is sent back through the network, modifying the weights automatically to minimize the actual and expected output difference. The current iteration's output becomes the input and influences the following iteration's output. This process is repeated until the desired output is obtained.

Each neuron performs the following computations in the output and hidden layers:

$$Out(x) = F \tag{1}$$

$$Hidden(x) = L(v(1) + M(1)x) \tag{2}$$

Here, $v(1)$ is the bias vectors, $M(1)$ is weight matrices, $F$ and $L$ are activation functions. The set of learning parameters $\theta = \{M(1), v(1)\}$. Typical choices for $L$ includes logistic sigmoid function with $sigmoid(b) = 1/(1 + e^{-b})$ tanh function with $tanh(b) = (e^b - e^{-b})/(e^b + e^{-b})$.

### 3.3. SVR model

SVR [27] classification is based on the same ideas of the SVM [28]. For predictive regression issues, it employs an Ɛ-insensitive loss function. This function finds a function that fits current training data with a deviation less or equal to Ɛ. The SVR algorithm transforms the original data points from the initial input space to a higher-dimensional feature space using a transformation function φ. A linear model is built in this new space, corresponding to a non-linear model in the old space. the feature space is identical to the input space where φ is the identity function, and the model generated is linear in the original space. The model employs a radius basis function (RBF) [29] kernel to transform data from non-linear to linear space. The kernel function enables the SVR to discover a good fit when data is mapped back to the original space.

### 3.4. Linear regression model

The linear regression model produces a diagonal straight line that depicts the relationship between variables. It returns the most optimal intercept and slope values. The dependent and independent variables are data features that remain unchanged. The intercept and slope are the values that influence the slope. There can be several straight lines depending on the intercept and slope parameters. The model works by fitting many lines on the data points and returning the line with the slightest error. It utilizes mean squared error (MSE) as a cost function to find suitable values that fit with data points [30].

### 3.5. Dataset preparation

COVID-19 data was collected from the district health information software (DHIS2) system [16] from March 1, 2020, to August 10, 2021. Daily reports and time-series sequential tables are included in datasets. The series table has been formatted in comma separated values (CSV) format. Confirmed and deaths were chosen for the attribute ordination. Before training importing libraries of Panda, NumPy, and Sklearn, the data was preprocessed following the steps mentioned in Figure 2.

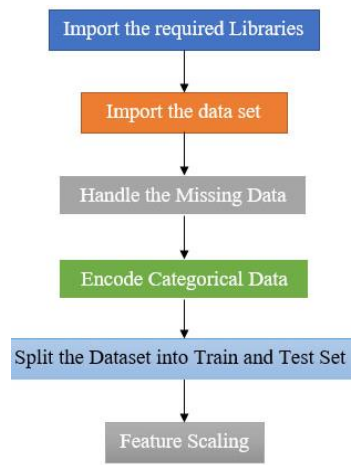

Figure 2. Dataset preprocessing

Tabular data was imported from the CSV file to create the feature's matrix and dependent vector with respective observations. Sklearn preprocessing library handled and encoded the missing and categorical data. The dataset was split into training, testing parts, and feature scaled using standard scaler of sklearn preprocessing. A visual representation of the dataset is shown in Figure 3. Figure 3(a) represents the number of confirmed cases and test positive rate concerning the number of tests, and Figure 3(b) shows the number of death and death rate scenarios.

From Figures 3(a) and 3(b), we have found that the coronavirus attacked Bangladesh in March 2020. The situation was under control until April 2021, but it increased. The country encountered the highest number of infected and death cases than previous between July 2021 to August 2021. Day-by-day data from March 2020 to August 2021 was collected to analyze and predict the situation. More discussion about the data has been mentioned in Table 1. The column 'number of data' and 'number of patients' represent the data collected for training. The next two columns, 'number of test data' and 'number of patients tested,' provides information about the data utilized for testing purpose. Data counting, mean deviation, standard divination, min, max, and quartile deviation are evaluated and enlisted in Table 1.

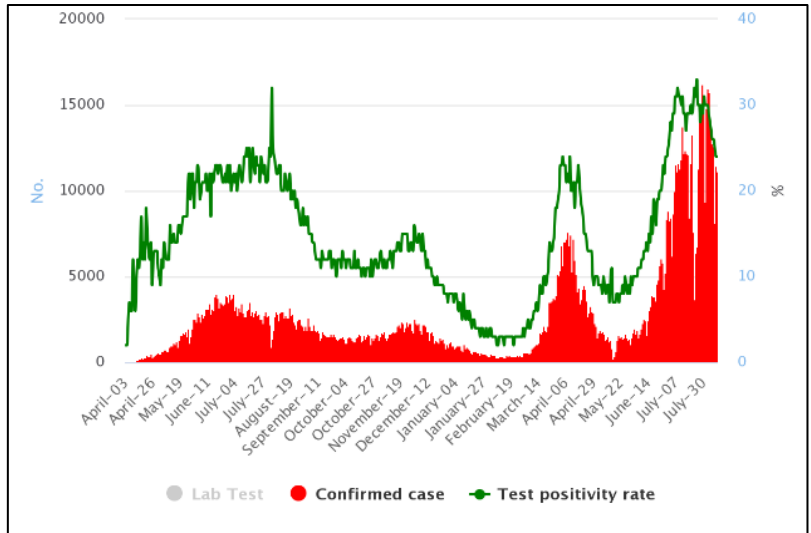

(a)

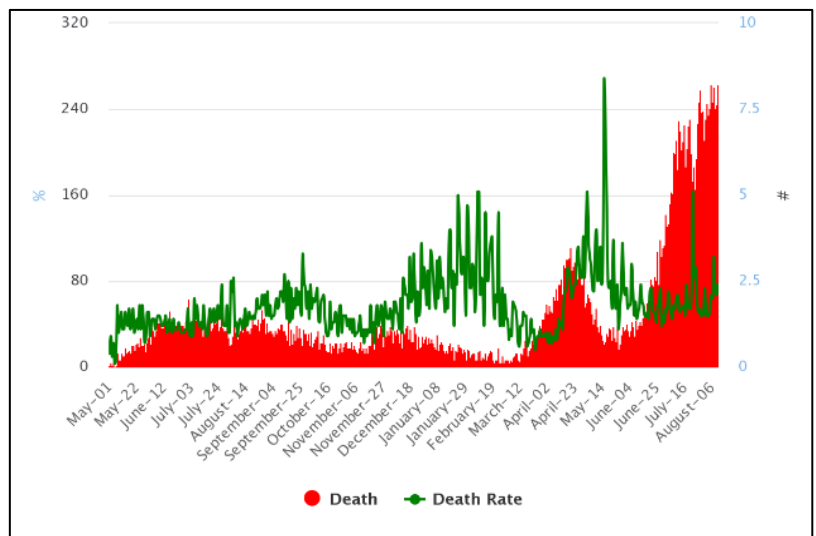

(b)

Figure 3. The number of cases (a) tests vs. confirmed and (b) death and death rate

Table 1. Dataset configuration

|  | Number of Data | Number of Patients | Number of Test Data | Number of Patients Tested |
|---|---|---|---|---|
| Count | 456 | 446 | 64 | 64 |
| Mean | 271.49 | 2894.19 | 272.25 | 2979.53 |
| Std | 152.29 | 3038.25 | 152.37 | 3320.52 |
| min | 1.00 | 5.00 | 7.00 | 112.00 |
| 25% | 142.75 | 1138.50 | 146.00 | 1197.75 |
| 50% | 272.50 | 1880.50 | 272.00 | 1836.00 |
| 75% | 402.25 | 3254.25 | 398.00 | 3151.50 |
| max | 533.00 | 16230.00 | 532.00 | 15776.00 |

**3.6. Hardware and software configuration**

A core i7 personal computer with 12 gigabyte (GB) random access memory (RAM), 4 GB Nvidia GeForce graphics memory, and 1 terabyte (TB) solid state drive (SSD) was used as the hardware components. As software components, the Linux operating system (Version 20.04.1-Ubuntu), Python 3.8.10, Panda 1.3.1, pip 21.3.1, scikit-learn 0.23.1, and the library of NumPy 1.19.5 were utilized for data processing, model training, testing, and analyzing purposes. To generate the plot, matplotlib 3.4.3 was imported.

### 3.7. Training and testing

The MLP, SVR, and linear regression models were trained, tested, and compared using the same dataset. Essential modules were imported to split the dataset into training and testing. 80% of data was taken for training purposes, and the rest 20% for testing the accuracy of the trained model. The independent train and test dataset were further scaled to ensure standard distributed zero centered and same order variance input data. The regression scores were evaluated from 5 Regressors to determine the best approach. For the SVR model, kernel RBF, poly degree=5, linear, poly degree=2, poly degree=7 were used to evaluate 1, 2, 3, 4, and 5 regressor's scores accordingly. For the MLP model, tanh activation function and lbfgs optimizer were commonly used in 1, 2, 3 regressors where maximum iterations were 1k, 5k, and 10k accordingly. In regressors 4 and 5, relu and tanh were activation functions, lbfgs and sgd were optimizers, and in both cases, maximum iterations were 1k. For the linear regression, learning rate were 0.5, 0.1, 0.01, 0.001, 0.0001 and number of iterations were 2.5k, 3k, 3.5k, 5k and 10k for 1, 2, 3, 4, 5 regressors accordingly. The scores generated by different models in different configurations have been listed in Table 2. Comparing the scores, the best model and approach were selected. Using the kernel linear, the SVR model achieved the highest score, and the linear regression model has gained the highest score using learning rate 0.5 and iteration 2.5k. The MLP model has generated the best regression score using tanh activation function and lbfgs optimizer, and it's the highest score than the other two models.

Table 2. Regression scores by different models

| Regressor No | MLP | | SVR | | Linear Regression | |
|---|---|---|---|---|---|---|
| | Infected Case | Death Case | Infected Case | Death Case | Infected Case | Death Case |
| 1 | 0.8669 | 0.8812 | 0.8230 | 0.8322 | 0.7996 | 0.8090 |
| 2 | 0.8758 | 0.8976 | 0.8413 | 0.8633 | 0.6954 | 0.7796 |
| 3 | 0.9182 | 0.9341 | 0.8373 | 0.8701 | 0.6405 | 0.6819 |
| 4 | 0.8772 | 0.8571 | 0.7322 | 0.7798 | 0.7363 | 0.7430 |
| 5 | 0.7632 | 0.7581 | 0.6909 | 0.7154 | 0.7073 | 0.6448 |

## 4. RESULT AND ANALYSIS
### 4.1. Infected and death prediction

To predict the counting number of COVID-19 infected and death cases of Bangladesh, the data from March 2020 to August 2021 was collected and processed. A machine learning model MLP was trained to analyze and generate a prediction. The model forecasts the infected and death rate between August 11, 2021, and September 11, 2021. Figure 4 represents the infected case. Figure 4(a) shows the exponential representation of the data where the estimated red line drawn by the MLP model has touched and passed through the nearest portion of most data values. Figure 4(b) represents a scatter plotting diagram highlighting the predicted and original test values. From the diagram, it can be concluded that most predicted values have passed through the nearest position of the original test value. The mean squared error for infected case prediction is 0.10475743200604047, where the model's regression score is appraised at 0.9182811910841436.

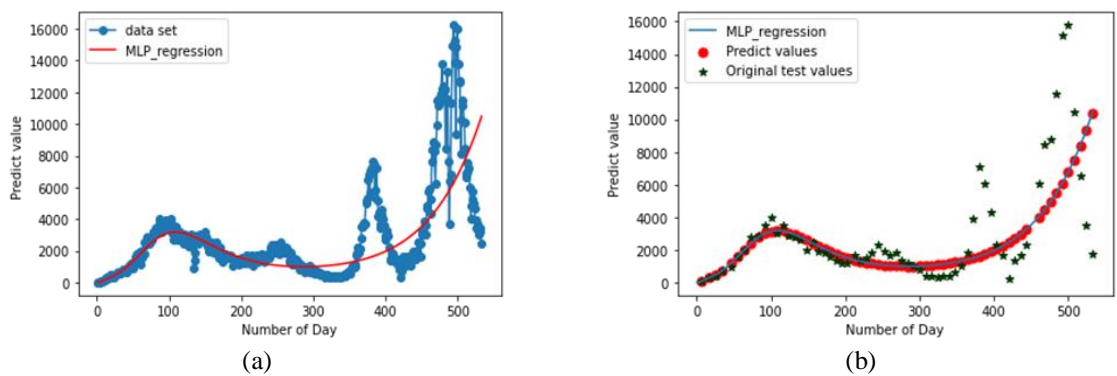

Figure 4. Infected case (a) exponential representation of data and (b) scatter plotting representation of data

Figure 5 shows similar information to the previous figure for the death cases. The estimated line in Figure 5(a) has touched and crossed from the nearest position of the most data values. In Figure 5(b), the predicted and original values are closely situated. From the calculation, 0.09692374417231872 is recorded as the mean squared error score, and the regression score is 0.9341569023686556.

Figure 6 exhibits the infected and death cases employing a line chart where the patterns at the end signify the prediction. The model presents forecasts for 30 days (August 11, 2021, to September 11, 2021). Figures 6(a) and 6(b) show curly lines representing the number of infected and death cases and their prediction. From the prediction, it can be noticed that for the next 30 days, the number of infected cases can be between 8,574 to 9,581, and death cases can be between 202 to 250. The actual number of cases was collected from the trusted government website within the prediction time and examined. The number of infected and death cases was close to the prediction by the model.

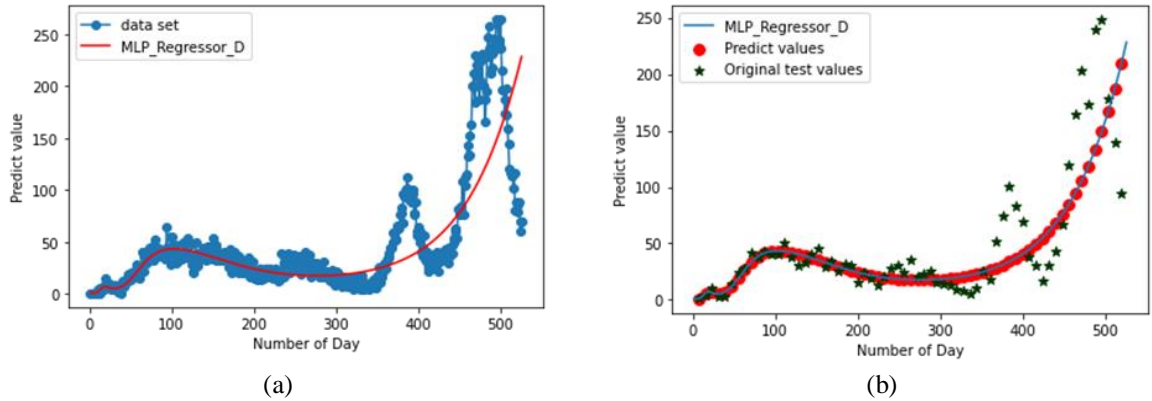

Figure 5. Death case (a) exponential representation of data and (b) scatter plotting representation of data

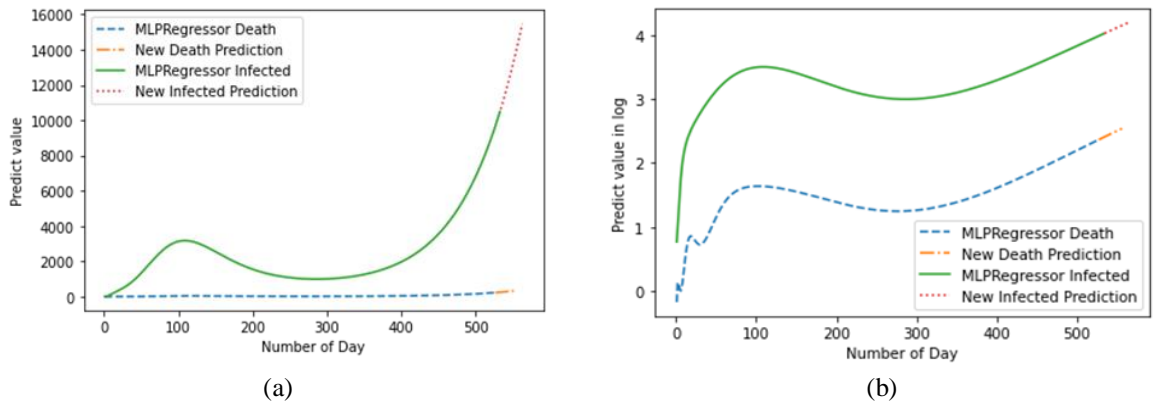

Figure 6. Infected and death case prediction (a) exponential representation of cases and (b) logarithmic representation of cases

**4.2. Critical situation analysis and prediction**

In Bangladesh, COVID-19 commenced affecting people in March 2020. From the starting time to mid-June 2021, the maximum number of infected and death were recorded at 400 and 113 accordingly. From mid-June 2021 to mid-August 2021, the delta variant attacked, and the number of infected and death cases reached the highest, and the country had never encountered such a massive number of cases before. In Bangladesh, the attack is famous as COVID-19's second wave. According to medical experts, other new varients can attack, and the situation may become critical. The MLP model has completed unique analysis and prediction to understand the upcoming third wave better. Figures 7 and 8 show the exponential and scatter plotting representation of infected and death cases data for the critical situation. From Figures 7(a), 7(b), 8(a) and 8(b), it is noticeable that the estimated line drawn by the model fits with most of the data values where predicted values and original test values are closely situated. The model achieved 0.0992140296903675, 0.0492014632878128 mean squared error and performed 0.9342422839997623, 0.950661069466337 regression score accordingly on infected and death case data.

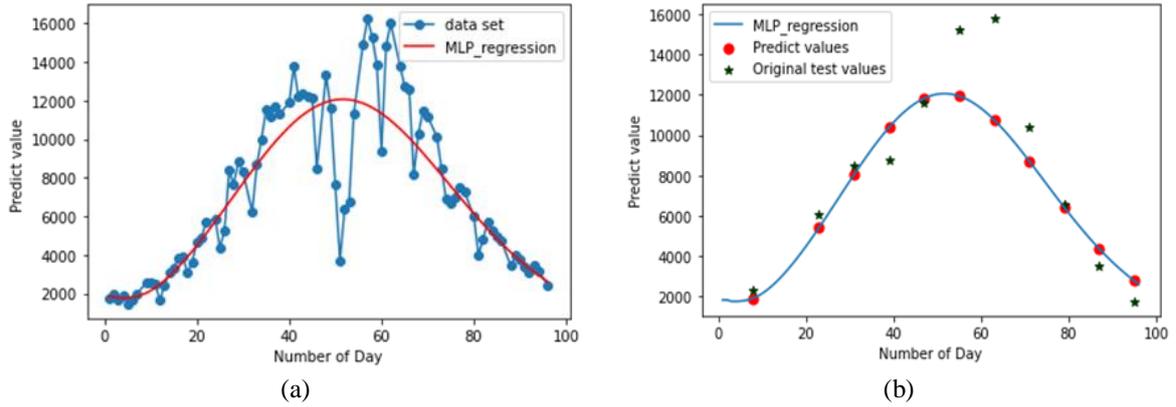

Figure 7. Infected cases in critical situation (a) exponential representation of data and (b) scatter plotting representation of data

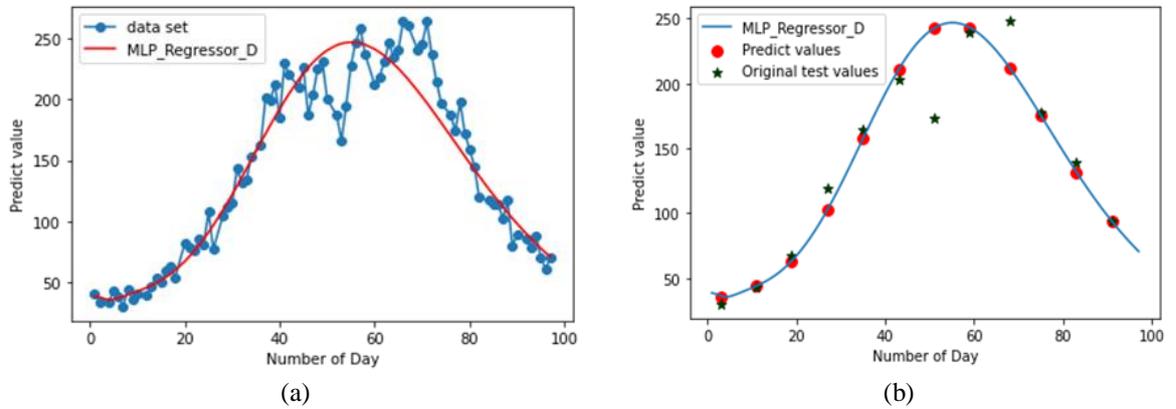

Figure 8. Death cases in critical situation (a) exponential representation of data and (b) scatter plotting representation of data

Figure 9 represents the infected and death case prediction in a critical situation where Figure 9(a) shows exponential, and Figure 9(b) shows a logarithmic representation of data. According to the model, the number of infected cases will be between 929 to 2,443 and death cases between 19 to 57. The infected and death rate will be minimized gradually.

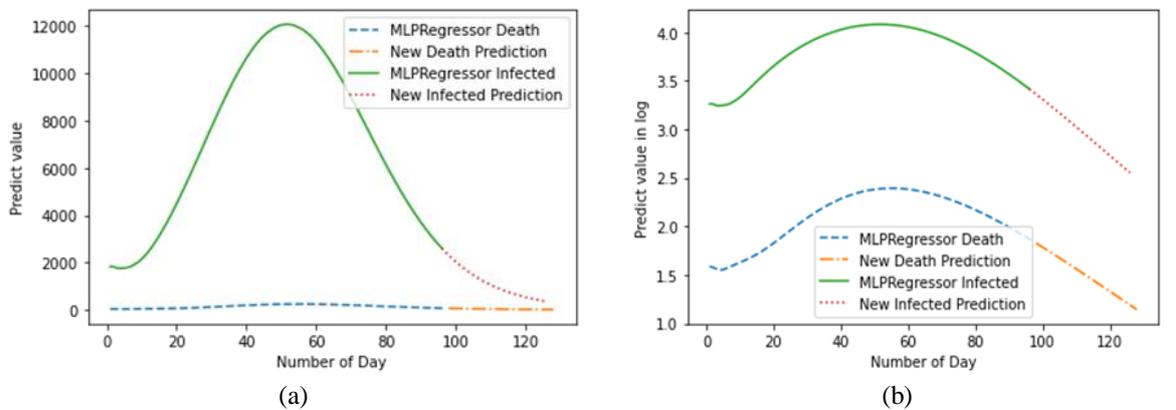

Figure 9. Infected and death case prediction in critical situation (a) exponential representation of cases and (b) logarithmic representation of cases

### 4.3. Prediction comparison

The MLP model was compared with two other well-known techniques, SVR and linear regression. For evaluating the performance of different techniques, the models were trained and tested with the same dataset. Figure 10 represents the prediction comparison among different models. Figure 10(a) shows infected, and Figure 10(b) shows death case prediction where the MLP model predicted more accurately. The prediction result was very close to the actual situation reported to the COVID-19 related Bangladesh government website.

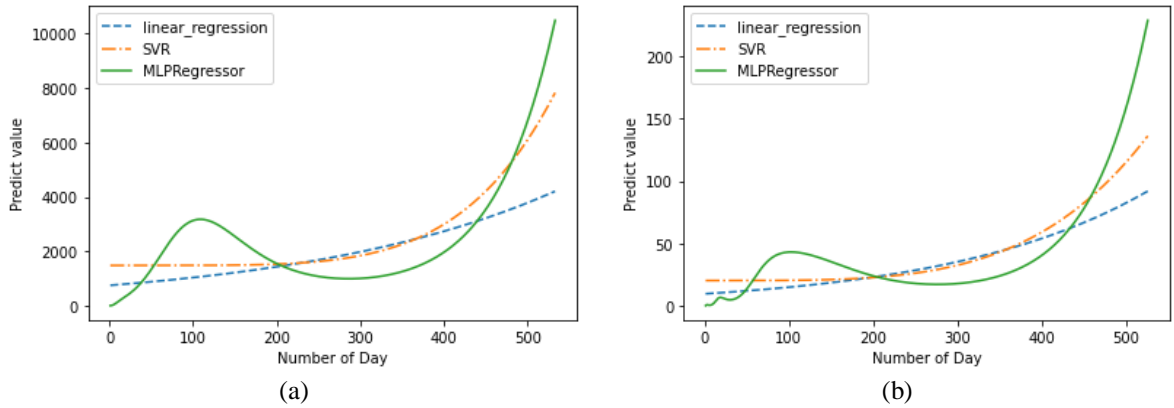

(a) (b)

Figure 10. Prediction comparison (a) infected case and (b) death case

The models performed another comparison for analyzing and predicting critical situations shown in Figure 11. The infected rate as shown in Figure 11(a) and death rate as shown in Figure 11(b) were compared using line charts. Comparing the performance of the MLP model, we expect that the prediction made by the MLP model will have a great match with the actual situation in the future.

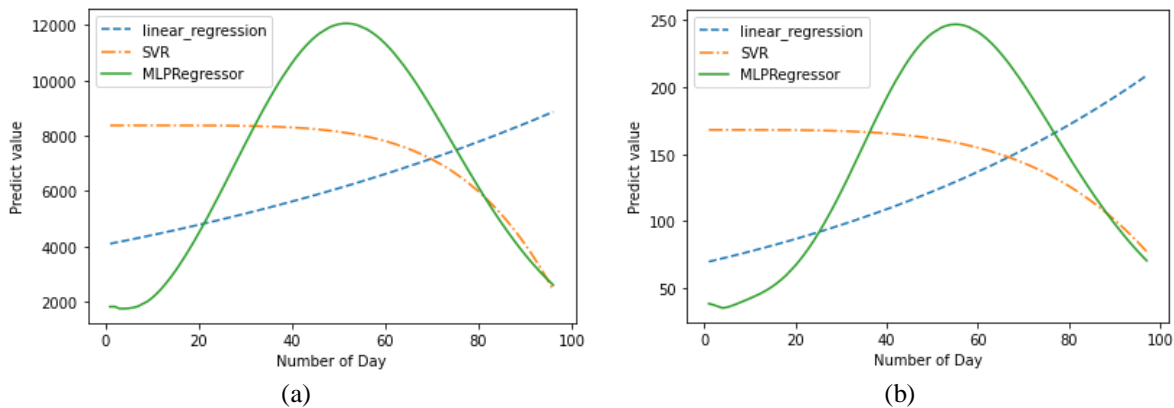

(a) (b)

Figure 11. Prediction comparison at critical situation (a) infected case and (b) death case

### 5. CONCLUSION

Over 2.3 million people have died because of the coronavirus's illness. In the year 2020, the global pandemic caused by this virus has impacted the lives of everyone, and the primary target has become to fight against it. Bangladesh's fight against COVID-19 is currently in one of its most critical stages because, during the last five months, the country has been severely impacted. machine learning can generate exact predictions about how the epidemic will affect us. We have tried to make advanced predictions through machine learning techniques to help the authorities understand the upcoming situation. Our model's accuracy and prediction capability were compared with state-of-the-art approaches, and we found satisfactory performance. The analysis and prediction results have a great match with the actual situation of Bangladesh. The results will benefit the authority to make strategic planning, resource allocation, and proper utilization, resulting in the least amount of human life loss. This work also serves as a foundation for future academics who want to develop better machine learning models for better data analysis. The accuracy of the model can be improved in various ways. In our research work, we were concerned about the impact of vaccination. Still, we noticed that the steps taken by the Bangladesh government have a significant impact on the minimization of death and infected rate. For future work, we plan to conduct another study considering the impact of vaccination.

## ACKNOWLEDGMENTS
The research is supported by Chulalongkorn University and Iwahori's research. Iwahori's research is supported by JSPS Grant-in-Aid for Scientific Research (C) (20K11873) and Chubu University Grant.

## BIOGRAPHIES OF AUTHORS

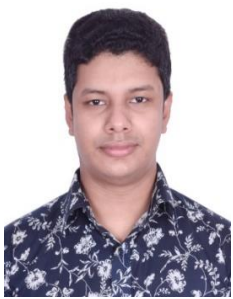

**Al-Akhir Nayan** received Bachelor of Science degree in Computer Science and Engineering from University of Liberal Arts Bangladesh (ULAB). He is enrolling as a master's student at Chulalongkorn University, Bangkok, Thailand. He does research in Machine Learning, Artificial Neural Network, and IoT. He can be contacted at email: asquiren@gmail.com.

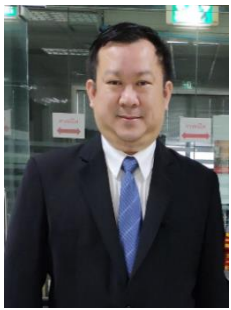

**Boonserm Kijsirikul** received Ph.D. from Tokyo Institute of Technology. He currently works at the Department of Computer Engineering, Chulalongkorn University, Bangkok, Thailand. He does research in Artificial Intelligence, Machine Learning, and Artificial Neural Network. He can be contacted at email: boonserm.k@chula.ac.th.

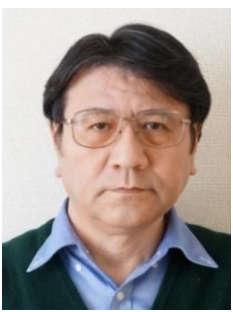

**Yuji Iwahori** received Ph.D. degree from Tokyo Institute of Technology. He joined Nagoya Institute of Technology and became a professor in 2002. He has joined Chubu University as a professor since 2004. He has been a visiting researcher of UBC since 1991, a research collaborator with IIT Guwahati since 2010, and Chulalongkorn University since 2014. He has been Honorary Faculty of IIT Guwahati since 2020. He does research in Computer Vision, Artificial Neural Network, and Machine Learning. He can be contacted at email: iwahori@isc.chubu.ac.jp.